# M3H: Multimodal Multitask Machine Learning for Healthcare

**Abstract**

Developing an integrated many-to-many framework leveraging multimodal data for multiple tasks is crucial to unifying healthcare applications ranging from diagnoses to operations. In resource-constrained hospital environments, a scalable and unified machine learning framework that improves previous forecast performances could improve hospital operations and save costs. We introduce M3H, an explainable Multimodal Multitask Machine Learning for Healthcare framework that consolidates learning from tabular, time-series, language, and vision data for supervised binary/multiclass classification, regression, and unsupervised clustering. It features a novel attention mechanism balancing self-exploitation (learning source-task), and cross-exploration (learning cross-tasks), and offers explainability through a proposed TIM score, shedding light on the dynamics of task learning interdependencies. M3H encompasses an unprecedented range of medical tasks and machine learning problem classes and consistently outperforms traditional single-task models by on average 11.6% across 40 disease diagnoses from 16 medical departments, three hospital operation forecasts, and one patient phenotyping task. The modular design of the framework ensures its generalizability in data processing, task definition, and rapid model prototyping, making it production ready for both clinical and operational healthcare settings, especially those in constrained environments.

## 1. Introduction

**1.1 Background**

The integration of Artificial Intelligence (AI) and Machine Learning (ML) has seen unprecedented promise to advance healthcare services and to fundamentally improve our understanding of medicine (Topol 2019, Yu 2018). Leveraging the increasingly accessible patient digital records, multimodal learning incorporates multiple modalities and sources of data input to provide holistic views of patient profiles (Soenksen 2022,

Huang 2020, Acosta 2022, Baltrusaitis 2019, Ahmed 2020). However, beyond the integration of diverse inputs, a combination of outcomes is often further necessary to characterize patients comprehensively. Multitask learning, which leads to performance breakthroughs in large language models such as GPT-2 (Radford 2019), and computer vision (Ren 2015, He 2017, Reed 2022), is a natural extension under this premise to simultaneously learn multiple medical tasks to improve model performance across cardiology (Torres-Soto 2020), psychiatry and psychology (Tseng 2020, Lee 2021), oncology (Fu 2021), radiology (Jin 2021) and other healthcare domains (Eyuboglu 2021, Wang 2023, Tang 2023). Specifically, in contrast with multiclass learning of mutually exclusive targets, multitask learning can simultaneously process multiple tasks and thus provide better performance due to the sharing of common knowledge. Importantly, multimodal multitasking emulates existing collaborative efforts in clinical settings, where physicians and administrators across multiple departments often integrate diverse sources of information to jointly navigate multiple complex medical decisions simultaneously. However, it remains challenging to develop an integrative multimodal multitask machine learning framework that is consistently applicable across distinct healthcare domains and machine learning problem classes while maintaining efficiency in handling increasingly large healthcare datasets (Ahmed 2023).

**1.2 Contributions**

M3H addresses several challenges, including the difficulty of integration across multiple distinct machine learning problem classes into a single framework and the lack of explainability metrics to measure how and why combining certain tasks improves performance. In particular, the M3H framework complements and extends previous literature on important key topics and provides new perspectives on the following:

1. M3H represents the first integrated healthcare system to bridge multi-disease diagnosis to hospital operations and patient phenotyping. In relation to this, it also represents the first step towards integrating not only clinical but also operational and biological dynamics of patient care, signaling a shift towards a holistic view across the healthcare continuum. M3H also facilitates integrating the most studied machine learning problem classes, including binary supervised classification, multiclass classification, and regression, as well as unsupervised clustering, thus offering unprecedented access and flexibility for both clinicians and healthcare practitioners to combine analytical tasks across the technical boundaries of the problem of interest, irrespective of how it was originally defined.

2. M3H introduces the TIM score, an explainable metric measuring incremental performance benefits from training additional tasks in conjunction with the source task. While previous studies in this field rely on apriori assumptions about the quantitative and qualitative value of multitask learning for the target domain or rely on medical observations, M3H provides a novel explainable mechanism that systematically quantifies the value of adding additional jointly learned tasks in a more rigorous level of granularity and broader combinatorial outcome space than any prior studies. Our extensive testing – which covers model experimentations across 40 disease diagnoses, three hospital operation tasks, and one patient phenotyping task – provides robust empirical evidence supporting our claim that multitask training can improve the performance of single-task models if the tasks are judiciously chosen. Conversely, our findings also reveal previously unreported phenomena where certain joint training diminishes the analytical gain of the canonical (single task) models. This observation implies the existence of competing objectives, which calls for the need to thoroughly explore the dynamics of task interactions, a process facilitated by the TIM score and its associated analyses.

3. M3H develops a novel cross-task attention mechanism that explicitly models the learning between medical tasks by balancing self-exploitation (learning for the source task) and cross-exploration (learning from other tasks). The unique attention structure is tailored for the understanding of complex interdependencies among multiple medical tasks as well as the facilitation of co-learning among tasks. Our attention mechanism also improves the overall ability to draw relevant and actionable insights with an attention weight, an explicit quantification of how much each task is contributing to the learning of another, enabling direct access to the decision process of multitask learning.

The rest of this paper is structured as follows. Section 2 outlines previous works that address multimodal multitask machine learning in healthcare settings. Section 3 details the architecture of the M3H framework including the novel attention mechanism. Section 4 describes the experimental setup using a large-scale intensive care unit (ICU) database. Section 5 demonstrates M3H's performance across a diverse set of medical and machine-learning tasks. The explainability metric is characterized in Section 6, and managerial implications, limitations, and future directions are discussed in Section 7.

## 2. Related Literature

### 2.1 Integrated Healthcare System

Medicine is not a standalone domain of study. On the quite opposite, medical departments rely heavily on the support and interactions of inter-departmental collaborations. Current studies in healthcare management primarily rely on a single-disease prediction for a single medical diagnosis, treatment, or planning problem, spanning from diabetes (Hajjar 2023, Kraus 2024), obesity (Yang 2013), milk bank planning (Chan 2023), to cardiovascular disease (Wang 2023). These domain-specific models offer expert insights into a particular domain of healthcare and could benefit significantly from sharing knowledge with each other if jointly studied under a unifying framework. Some recent works on the integration of logistical scheduling and staffing operations across radiology (Bentayeb 2023) and nurse management (Kim 2015) as well as validation of improved quality and reduced re-admission rates under collaborative healthcare structures (Lan 2021), all show promising potential of an integrated healthcare system for improved operational efficiencies and qualities. In addition to medical task integration, methods have also been developed to unify analytical tools in an integrated predictive and prescriptive modeling approach (Bergman 2020).

A further testimony to this critical direction towards integration is the rising interest in the medical field in the study of multi-disease (Buergel 2022), or multi-morbidity diagnosis (Zhao 2024). Such approaches make heavy use of the underlying assumption that patient characteristics, as well as medical conditions, when studied holistically, provide a better clinical understanding and thus improve both model performance as well as analytical insights. Furthermore, these studies imply significant managerial benefits for the patients, caregivers, and even payers (Bish 2024, Apergi 2023). A concrete illustration was demonstrated by the prediction of patient flow in a large hospital system (Bertsimas 2022, Bertsimas 2023, Na 2023), where predictions of multiple operational targets, including length of stay, mortality, and ICU admissions, are all studied to characterize the patient's condition. Under the M3H framework, hospital systems could benefit from clinical and operational forecast performance improvements, directly contributing to operational efficiency, and profits for the organization.

## 2.2 Explainability of the Outcome Space

Existing explainable multitask frameworks for learning biomedical tasks focus on explaining the contribution of input features on its corresponding outcome task (Tang 2023). However, multitask learning offers a unique perspective that lies in the interactions among the jointly learned tasks. The majority of past multitask works in healthcare rely on prior medical knowledge that learning specific task combinations is beneficial to each problem (Wang 2023, Tang 2023). This approach heavily relies on expert understandings of the domain studied and introduces an unmet need for a rigorous procedure to select and combine tasks. Such task-interaction understanding is critical both clinically and managerially. Although no previous literature has explicitly reported the direct connection, the authors conjecture that it is possible that multitask learning could provide a new angle for the automatic learning of seemingly counter-intuitive medical domain interactions. Recent medical works have discovered hidden links in several previously unexplored human "axes", including the gut-brain axis, connecting gut microbiome and Alzheimer's disease (Aura 2023). These works potentially imply that beyond the interactions of the input space (features), interactions of the output space (outcomes), are the next step for our understanding of medical systems.

## 2.3 Integration Across Machine Learning Classes

Early work in multitask learning demonstrated superior performance in source tasks when combined with additional tasks in translation (Radford 2019), and object detection (Ren 2015). Later works have devoted significant efforts in trying to integrate problems of different domains into a single framework via the use of aggregated loss function (Wang 2023), cross-stitch networks (Ishan 2016), and attention mechanism (Liu 2018). Following this body of study, we propose to integrate the four studied machine learning problem classes via an aggregated loss function and by proposing a novel attention mechanism to study the cross-task interactions between learned tasks. A notable difference between this work and the previous is the attention mechanism' design for future interpretability purposes: specifically, we design the first known architecture that projects each medical task into a universal embedding and explicitly controls for self-exploitation and cross-explorations. These design features allow users to visualize post-model-fitting how tasks are learned during training, providing access to interpret the learning mechanism.

## 3. M3H Architecture

### 3.1 Overview of the M3H framework

M3H (Fig. 1.) is developed as a generalizable and explainable framework that leverages multimodal multitask machine learning to improve medical task performance and task-dependency understanding. Qualitatively, M3H improves previous works in this field (Torres-Soto 2020, Tseng 2020, Lee 2021, Fu 2021, Jin 2021, Eyuboglu 2021, Wang 2023, Tang 2023) by providing an integrated pipeline across a diverse set of medical tasks (*e.g.* diagnosis, operations, and phenotyping) and technical tasks (*e.g.* binary classification, multiclass classification, regression, and clustering), and provides explainability not only on the contribution of input features but on the output task interactions. It further proposes a novel attention mechanism to facilitate cross-task learning.

The M3H framework is an end-to-end framework for integrating multimodal data feature extraction and multitask outcome learnings. To leverage the strong performance of existing state-of-the-art (SOTA) models, M3H first obtains fixed modality-specific embeddings through publicly available, pre-trained models, including ClincalBERT (Alsentzer 2019) for natural language and Densenet121-res224-chex (Cohen 2021) for images. These task-agnostic, not-trainable multimodal embeddings are then passed through further modality-specific learnable feedforward networks and then integrated into a shared-task learning module. In this module, we conduct (i) contrastive learning and (ii) shared-task learning, where the first aims to project embeddings from different modalities into a consistent embedding space, and the second serves as an over-arching tunnel that all tasks must contribute to learning and a proxy for a universal embedding that is relevant for all tasks. We then feed the shared-learned embedding to task-specific networks, which focuses on the learning of each individual task. Finally, these task-specific embeddings integrate knowledge from other task embeddings before making their final predictions via the cross-task attention mechanism.

In multimodal multitask machine learning problems, it remains challenging to unify a diverse pool of outcomes due to the presence of different output spaces (continuous numeric, discrete categories). M3H integrates tasks of different medical domains and machine learning problem classes by unifying losses from each sub-problem into a single objective function. Specifically, the overall loss is a combination of contrastive loss between multimodal inputs and aggregated problem class loss of jointly learned outcomes. During training, network updates are made by optimizing each individual loss sequentially.

### 3.2. Machine Learning Problem Class

We detail below the four machine learning problem classes that we consider in the framework, why they are important cases to consider, and their evaluation metrics. They are separated into two categories: supervised (observable true labels) and unsupervised learning (no observable true labels).

***Supervised-Binary Classification:*** Binary classification is the most widely used form of problem class in healthcare and refers to the prediction of two classes of outcomes: positive and negative. In settings where only 2 discrete outcomes are present (i.e., infected or not), or where a clear numerical cutoff separates two conditions (i.e., death after 48 hours), we use binary classification to understand the variable relationship and evaluate the performance using the area under the receiver operating curve (AUROC). Specifically, for tasks with overly imbalanced classes (positive sample less than 10% of cohort), we initialize the output layer's bias as $\log(n_{positive}/n_{negative})$ to de-bias the imbalance following previous literature (Abadi 2015).

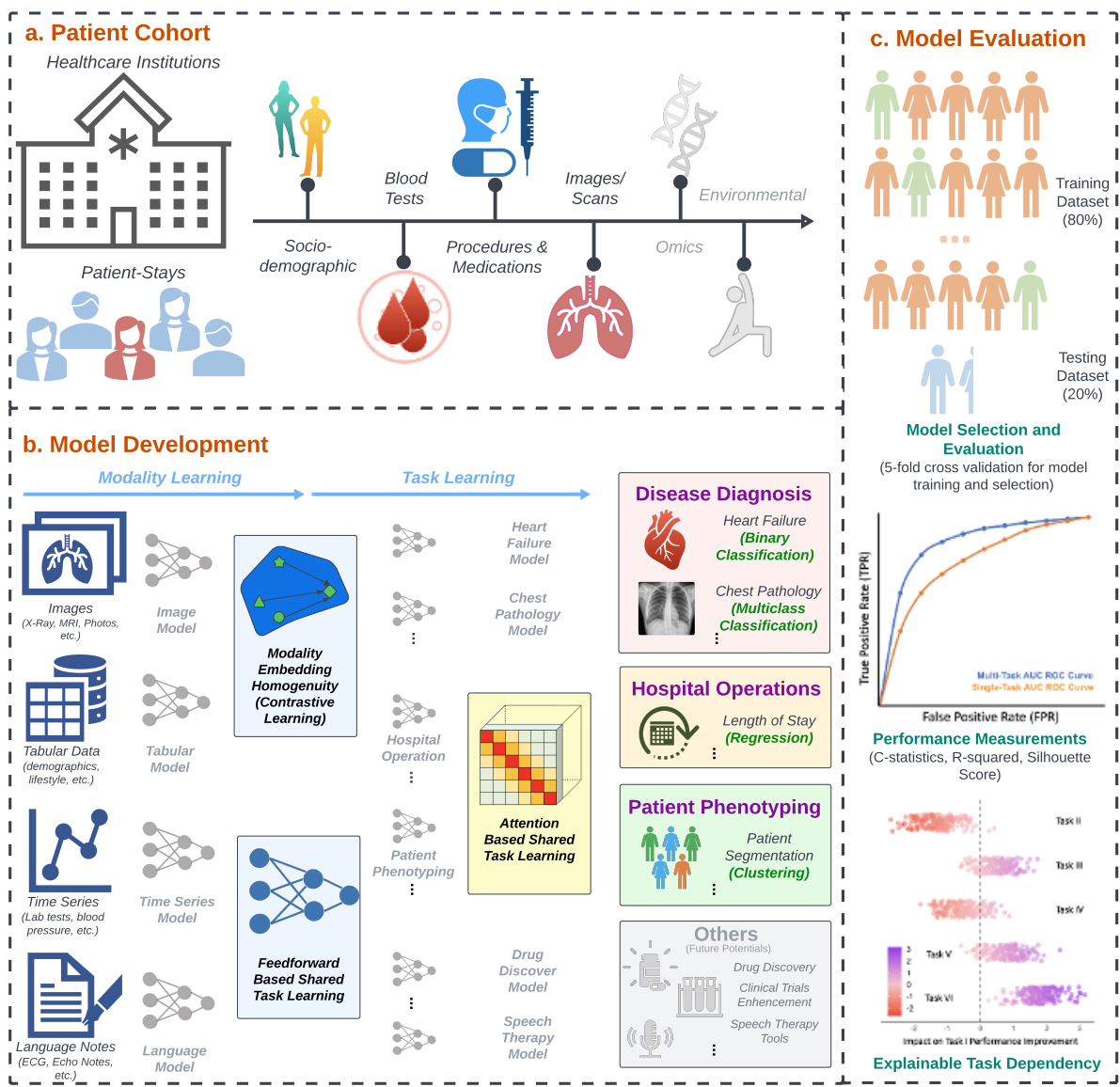

**Figure 1. Multimodal Multitask Machine Learning for Healthcare (M3H) system.**

***Supervised-Multiclass Classification:*** Multiclass classification refers to the prediction of multiple mutually exclusive classes of related outcomes and is studied when inherent structures between classes are present and related, including device-specific identifications (Anguita 2012) and anatomic-specific disease detections (Pogorelov 2017, Alshmrani 2022, Hoang 2022). Due to the natural imbalance of certain class outcomes, we evaluate it by computing the averaged AUROC across all classes.

***Supervised-Regression:*** Regression problem refers to the predictive modeling of continuous numerical outcomes. It is especially important when the prediction of the exact values is cost-sensitive, which has a large impact on care access fairness (Reid 2008), operational efficiency (Caruana 2015), and physician wellness (Salari 2020). R-squared value is chosen as the evaluation metric for regression problems.

***Unsupervised-Clustering:*** Clustering problems are fundamental in exploratory efforts to understand data structures without applying pre-determined labels for training (Haraty 2015), as well as anomaly detection for healthcare fraud prevention (Hillerman 2017). A high-quality cluster assignment achieves a high silhouette score, an evaluation of both separability, the ability to distinguish clusters from one another, and homogeneity, the ability to have consistent content within each cluster.

*3.3 Machine Learning Problem Class Architectures*

The M3H framework assigns a pre-defined modality-specific feedforward network for each input modality and a task-specific network for each outcome task, with details of each network in Supplemental Fig 1. and the overall pipeline in Supplemental Fig. 2. Specifically, as an unsupervised problem with unlabeled samples, clustering is uniquely challenging to incorporate into the M3H framework. Unlike the use of output layers to predict a predefined label, to effectively group patients into different phenotypes, we train an autoencoder that learns accurate low-dimensional latent space that can be then clustered into groups via traditional methods such as K-means clustering. The autoencoder architecture can be found in the Supplemental Fig. 1. We first concatenate all embeddings from all modalities into an aggregated embedding, this embedding is then fed during training into an autoencoder to compress the original feature space into low-dimension latent space and then re-expanded back to the original dimensions. A good quality latent space aims to achieve low reconstruction loss, measuring a low difference between the encoder input and the decoder output of the autoencoder. The learned latent space is then clustered into 15 patient subgroups and evaluated for quality, as illustrated in Fig. 2a.

*3.4 Cross-Task Attention for Knowledge Sharing*

Attention mechanism (Vaswani 2017) is at the foundation of recent breakthroughs in artificial intelligence. At its core, attention is constructed by variations of the key, query, and value vectors to capture interactions between its inputs. Attention mechanism is well-positioned to exploit dependencies between tasks: by projecting each task's embeddings as a token, we can leverage the attention mechanism to enable explicit task knowledge sharing. The overall architecture is demonstrated in Fig. 2b.

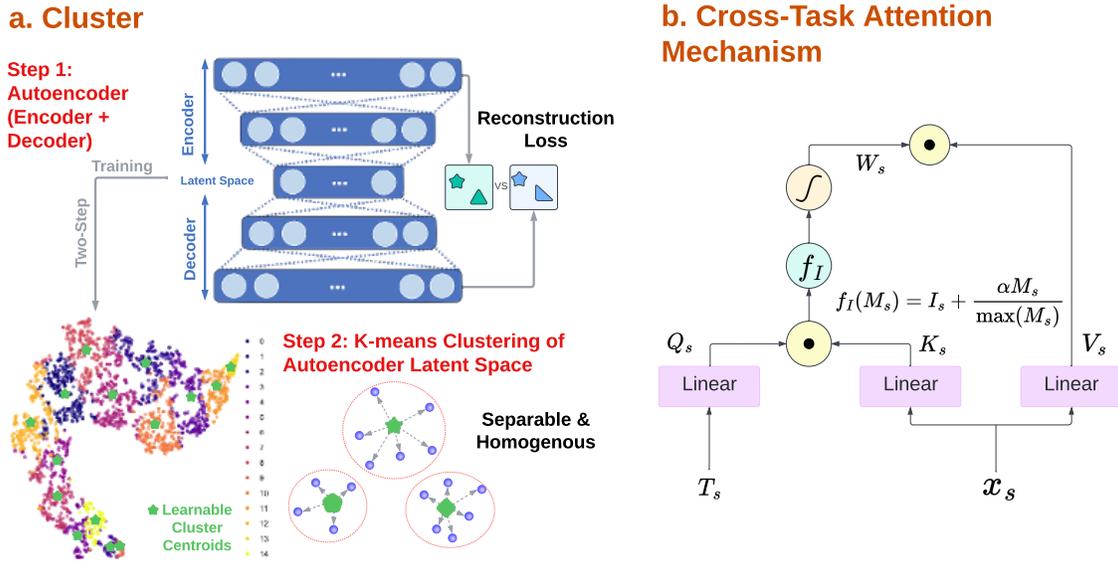

Figure 2: Architecture Design of the Clustering and Cross-Task Attention Mechanism.

Following traditional practice, input to obtain the key and value vectors is the original task embedding $x_s \in \mathbb{R}^{n_{batch} \times n_{tasks} \times n_{feature}}$ from joint learning of a specified set of task $s$, where $n_{batch}$ refers to the number of samples passed through each learning iteration (batch size), $n_{feature}$ refers to the number of features generated to encode knowledge for each task, and $n_{tasks}$ refers to the number of tasks in $s$. However, we aim to find a universal mapping between task tokens indicating the index of a task, with the embedding that best represents a task. This calls for a query vector that is independent of the batch update. To do so, we generate the query embedding via mapping of the task tokens vector $T_s$ to task embeddings via a linear projection $Q_s = f(T_s): \mathbb{N}^{n_{tasks}} \to \mathbb{R}^{n_{tasks} \times n_{feature}}$. Finally, we apply linear projections to all embeddings to improve representation quality and obtain query vector $Q_s \in \mathbb{R}^{n_{tasks} \times n_{feature}}$, key vector $K_s \in \mathbb{R}^{n_{batch} \times n_{tasks} \times n_{feature}}$, and value vector $V_s \in \mathbb{R}^{n_{batch} \times n_{tasks} \times n_{feature}}$.

The product of the computed query and key vectors, referred to as attention weight, indicates the relevance or emphasis put on a specific token in the value vector. We aim to find a balance between exploiting self-learning (reusing knowledge from the original task) while exploring cross-learning (incorporating knowledge from other unrelated tasks) in a controlled manner. This balance is achieved by adapting the initial attention weight $M_s \in \mathbb{R}^{n_{batch} \times n_{tasks} \times n_{tasks}}$ through the projection $W_s = softmax(I_s + \frac{\alpha M_s}{\max(M_s)})$, where $I_s \in \mathbb{R}^{n_{batch} \times n_{tasks} \times n_{tasks}}$ is the identity matrix encouraging self-learning, and $\alpha$ is the strength of exploration encouraging cross task learning. This attention weight is then applied to the values vector to obtain the final cross-learned task embeddings. A detailed algorithm outline of the novel cross-task attention computation can be found in Fig. 3.

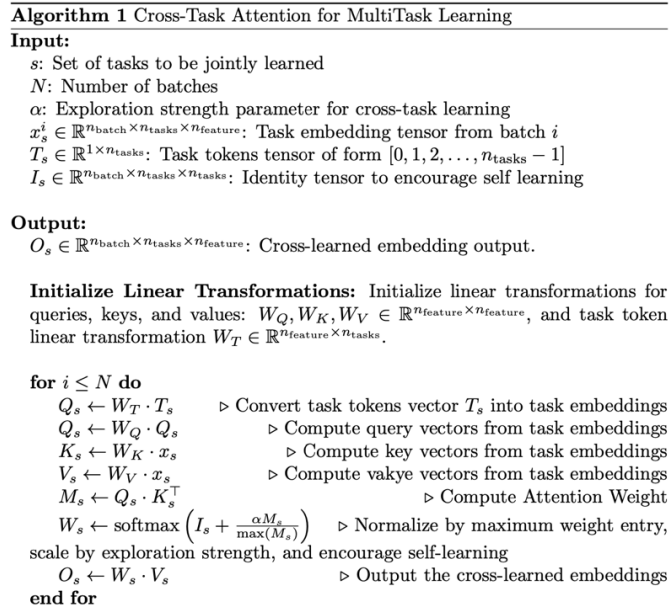

Figure 3. Algorithmic representation of cross-task attention learning.

## 3.5 Model Loss

The aggregated loss used to train to network is defined as a weighted average across all losses: $l_{total} = w_c \, l_{constrative} + \sum_{t \in B_s} w_t \, l_{binary,s} + \sum_{t \in M_s} w_t \, l_{multiclass} + \sum_{t \in R_s} w_t \, l_{regression} + \sum_{t \in C_s} w_t \, l_{cluster}$ where $w_c, w_t$ refers to the weight assigned for contrastive loss and task $t$, $B_s, M_s, R_s, C_s$ refers to the sets of tasks that are in the binary, multiclass, regression and cluster problem classes. Specifically, binary classification loss refers to binary cross entropy with logits loss; multiclass classification loss refers to negative log-likelihood loss; regression loss refers to mean absolute error loss; cluster reconstruction loss refers to the mean squared error between encoder input and decoder output. All weights have been initialized to 1 for ease of implementation.

## 4. Experiment Setup on Large-scale Medical Database

### 4.1. Dataset and Patient Representation

HAIM-MIMIC-MM (Soenksen 2022) is a patient-centric dataset derived from Medical Information Mart for Intensive Care IV (MIMIC-IV) (Johnson 2023), a public electronic health record database from Beth Israel Deaconess Medical Center containing de-identified records of all patients admitted to the intensive care unit (ICU) between 2008-2019. HAIM-MIMIC-MM offers access to contemporary, large-scale patient cohorts with modular constituent data organization, and most importantly, integrates multiple modalities of data inputs into a single database, ranging from demographics, chart events, laboratory events, procedure events, radiological notes, electrocardiogram notes, echo-cardiogram notes, as well as chest X-ray images. Specifically, HAIM-MIMIC-MM aggregates all available medical information of a patient's hospital admission-stay gathered before their expiration or discharge time, with their details summarized in Table 1. The dataset contains a total of 12,586 samples spanning 5178 patient hospital stays aged 18 – 91. The dataset integrates 4 distinct types of input modalities (tabular, time-series, language, and vision) and 11 data sources (for example, echocardiogram notes). Fixed-size vector representation (embeddings) of data from four modalities: tabular (dimension of 6), time-series (dimension of 451), vision (dimension of 2084), and language (dimension of 2304), are extracted using pre-trained, state-of-the-arts models and combined into a comprehensive multimodal patient representation. Each sample within the HAIM-MIMIC-MM dataset corresponds to all prior patient information from the time of admission until an inference event, including the time of imaging procedure for pathology diagnosis, the 48-hour window for mortality prediction, or the end of hospital stay. The rich patient profile developed in HAIM-MIMIC-MM allows for standardized, fast prototyping of relevant healthcare models.

| Modality | Source | # of Features |
|---|---|---|
| **Tabular** | Demographics | 6 |
| **Time Series** | Chart Event | 99 |
| | Lab Events | 242 |
| | Procedure Events | 110 |
| **Language** | Radiological Notes | 768 |
| | Electrocardiogram Notes | 768 |
| | Echo-cardiogram Notes | 768 |
| **Vision** | Chest X-ray images | 2304 |

**Table 1. Feature characteristics of the HAIM-MIMIC-MM database.** HAIM-MIMIC-MM is a comprehensive multimodal dataset that contains both the MIMIC-IV and MIMIC Chest X-ray images dataset that only includes patients that have at least one chest X-ray performed. It includes a diverse and rich pool of features that characterize the patient profile.

### 4.2 Medical Tasks of Interest

*Disease Diagnosis:* Early prediction of disease diagnosis is one of the most important and studied areas of healthcare due to its crucial role in enabling timely intervention to reduce severe complications and improve treatment success. It is particularly important to note an increasing interest across medical fields to try to understand connections between diseases across different medical domains, ranging from the discovery of the connection between gut microbiome and Alzheimer's to diabetes and heart diseases (Aura 2023, Rawshani 2018). In this study, we identified 40 commonly known conditions and diseases and grouped them into 16 clinical departments based on the Mayo Clinic disease and condition directory.

Specifically, we cover departments ranging from blood disorders, cardiology, critical care, dermatology, endocrinology, gastroenterology and hepatology, infectious diseases, internal medicine, nephrology, neurology, oncology, ophthalmology, psychiatry and psychology, pulmonology, rheumatology, and urology. Identification of each specific disease diagnosis is summarized in Supplemental Table 1, and we structure the problem as binary classification: identified ICD code is recorded in the patient's end-of-hospital discharge diagnosis record (1), otherwise (0). We also consider chest pathology diagnosis, which is structured as a multiclass problem with 8 common thoracic diseases: atelectasis, cardiomegaly, consolidation, edema, enlarged cardio mediastinum (enlarged CM), lung opacity, pneumonia, pneumothorax. Patients are included only if they have a single and unique pathology identified as positive, to avoid overlapping between outcome classes. The pathologies are selected so that no pathology occupies less than 1% of the entire patient population or contributes to a significant reduction of data sample size.

*Hospital Operations:* The daily operations of healthcare facilities have profound implications on their resource allocation, patient satisfaction, as well as clinician decision effectiveness and timeliness. Specifically, patient flow forecast (length of stay prediction) allows the logistics team to get advanced notice of discharge, ultimately improves patient quality of care and reduces operational costs in the hospital system; deterioration warnings (mortality in the next 48 hours prediction), especially in time-sensitive environments such as ICU units, guides physicians to make rapid, yet critical decisions in a data-intensive environment (Na 2023); safety enhancement (hospital-acquired infection prediction) prevents infection outbreaks with targeted control practices to safeguard patient health, incentivizing facilities to follow hygiene and sterilization protocols to avoid operational costs (Peasah 2013). We structure the length of stay prediction as a regression problem (unit of days), mortality prediction, and HAI prediction as binary classification problems. Specifically, following previous literature (Ishigami 2018), we identify catheter-related bloodstream infections, nosocomial pneumonia, surgical site infections, and urinary tract infections as HAI infections.

***Patient Phenotyping:*** As medical records become more standardized, recorded data may be structured in ways that are best suited for reimbursement and billing purposes. Such imposed bias could potentially miss clinical information that misclassifies patients into irrelevant patient subgroups. Patient phenotyping provides an important perspective into understanding additional structures shared by patient subgroups not traditionally defined. In our work, we look for the best patient grouping mechanism that groups the cohort into 15 phenotypes via clustering.

## 4.3 Model Training Pipeline

We initially explored various feedforward architectures for each modality-specific and task-specific network including different activation functions (ReLU, Sigmoid, Tanh), dropout layers, normalization layers, different optimizers (RMSprop, SGD, Momentum, Adam), and gradient clipping. The canonical architecture used in all following experiments was selected to support GPU optimization for computational efficiency (i.e., the number of filters in layers mostly are multiples of 64) and was shown to have a consistently superior performance during preliminary investigations. The rescaling coefficient in the cross-task attention mechanism $\alpha$ is set to be 0.1 as it is explored to be a stable point between performance stability and efficient learning.

We first split the dataset into 80% training (n=10025) and 20% testing (n=2561) by stratifying on a patient level to ensure no data leakage between training and testing for all model training or validation processes. We then apply a 5-fold cross-validation on the training set to select the best combinations of batch size (256, 512) and learning rate (0.0001, 0.0003). Specifically, within each run of 15 epochs, 4 of the 5 folds are used for model training, and the remaining one is used for validation. The average of all 5 validation scores across all tasks is computed for each hyperparameter combination, and a final model is trained on the entire training set with the hyperparameter with the highest average validation score.

As the number of tasks included in joint learning grows, there is an exponentially growing number of potential possible task pair combinations. We consider the following task selection procedure to optimize our likelihood of locating the best-performing multitask model: given a set of tasks $s \cup \{i\}$ and its performance on $i$, we conduct experiments on all possible $s \cup \{i\} \cup \{j\}$ where $j$ is a task not previously included. We only keep the best-performing top 3 pairs and repeat until no further improvements are observed. For computational efficiency, we restrict pairs experiments up to pairs of 3. The final combination is the optimal task combination that can be used to improve task performance for a specific task.

## 5. Experimental Results

### 5.1 Quantitative Performance Improvements Across Medical Tasks

We demonstrate the feasibility of the proposed M3H framework through its application to a pre-established and validated multimodal dataset. Across 16 disease groupings with 40 disease diagnoses, 3 hospital operations tasks (length of stay, general mortality, and hospital-acquired infection), and 1 patient phenotyping task, the M3H framework demonstrates consistent performance improvement over single-task models as demonstrated in Fig. 4. We report the percentage of improvement and its lower and upper bound accounting for standard deviation after applying bootstrapping on the out-of-sample test scores between the best-performing single-task models and best-performing multi-task models. Specifically, confidence intervals are obtained by computing the standard deviation (SD) of bootstrapping 90% of the test set 5 times (lower bound: test-set-point-estimate – SD, upper bound: test-set-point-estimate + SD). Performance scores are presented as point estimates on the test set, marked as purple squares, with confidence intervals (CI) generated from bootstrapping, marked with yellow horizontal lines. Multi-task models are also shown to have reduced variability with more narrow confidence intervals, implying their potential to generate more robust solutions on unseen datasets.

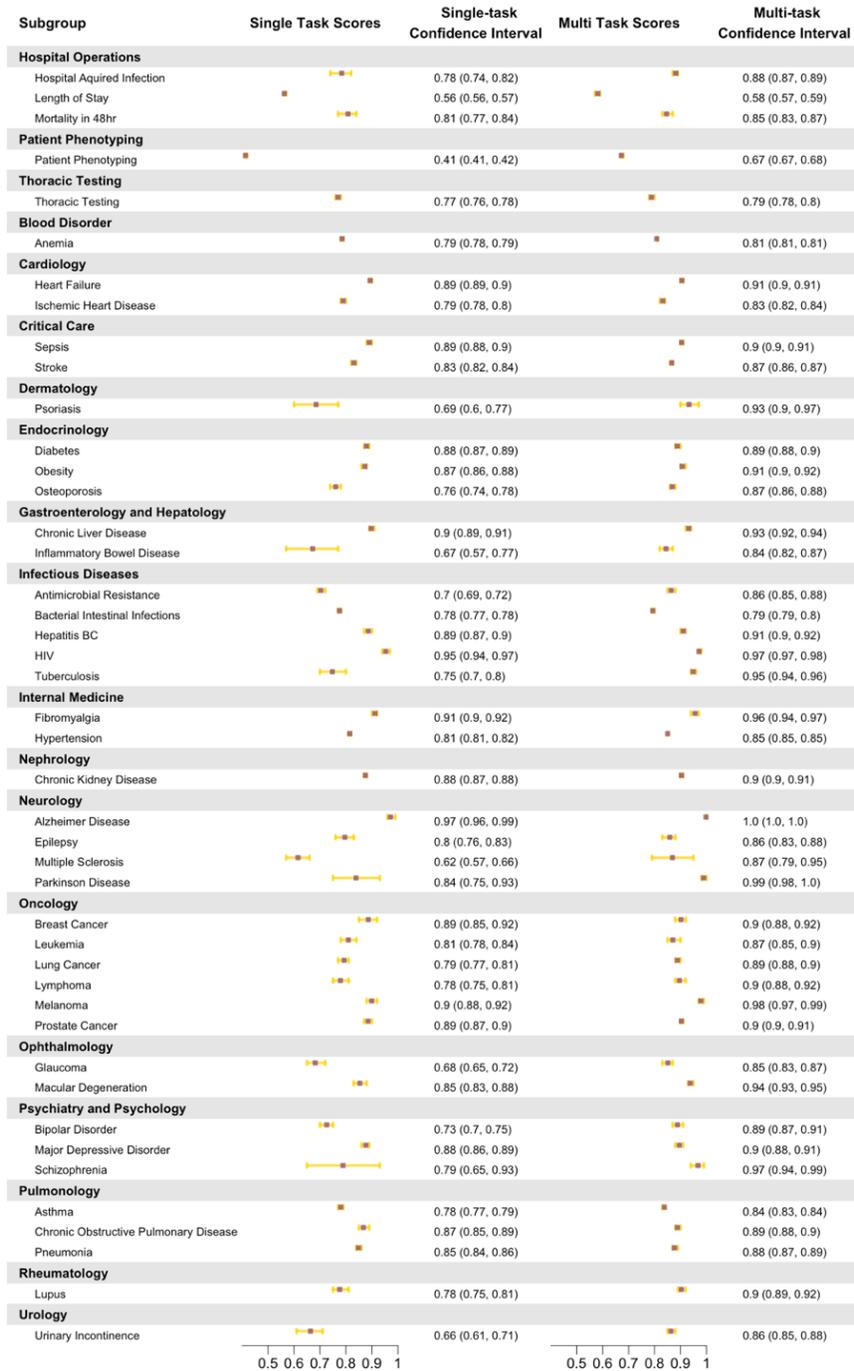

**Figure 4. Single-task vs. Multitask Models Across Important Healthcare Tasks.**

The M3H framework improves performance scores in diagnosis (1% – 41.2%), in hospital operations (3.3% – 12.4%), and in patient phenotyping (62.7%). Specifically, the improvement across disease groupings or hospital functionalities include hospital operations ($\Delta_{AUROC}$ = 4.7 – 12.4%, $\Delta_{R-squared}$ = 3.3%), thoracic testing ($\Delta_{Average\ AUROC}$ = 2.4%), blood disorder ($\Delta_{AUROC}$ = 2.9%), cardiology ($\Delta_{AUROC}$ = 1.4 – 5.3%), critical care ($\Delta_{AUROC}$ = 1.6 – 4.3%), dermatology ($\Delta_{AUROC}$ = 36.2%), endocrinology ($\Delta_{AUROC}$ = 1 – 14.1%), gastroenterology and hepatology ($\Delta_{AUROC}$ = 3.8 –25.7%), infectious diseases ($\Delta_{AUROC}$ = 1.8 – 26.9%), internal medicine ($\Delta_{AUROC}$ = 4.4 – 4.8%), nephrology ($\Delta_{AUROC}$ = 3.2%), neurology ($\Delta_{AUROC}$ = 2.7 – 41.2%), oncology ($\Delta_{AUROC}$ = 1.7 – 15.0%), ophthalmology ($\Delta_{AUROC}$ = 9.9 – 24.9%), psychiatry and psychology ($\Delta_{AUROC}$ = 2.1 – 22.7%), pulmonology ($\Delta_{AUROC}$ = 2.4 – 7.4%), rheumatology ($\Delta_{AUROC}$ = 16.4%), and urology ($\Delta_{AUROC}$ = 30.1%).

### 5.2 Generalizability across Machine Learning Problem Classes

Supervised and unsupervised machine learning (ML) have unique modeling techniques tailored for each corresponding outcome and objective. However, these distinctions of machine learning problem class should not pose barriers to integrating relevant tasks that can benefit from learning simultaneously. In M3H, we unify the learning of the most commonly used problem classes in healthcare: binary classification, multiclass classification, regression, and clustering, into a single framework. For binary classification, predicted major depressive disorder risk scores quantile, when compared against observed event rate, shows more consistency between female and male subgroups under the multitask setting, especially for low-risk patients (Fig. 5a); for multiclass classification, we observe reduced variability of ROC curve across different thorax conditions with higher averaged AUROC measure (Fig. 5b); for regression, multitask captures tail-predictions (extended length of stay) more closely than single-task (Fig. 5c); for clustering, post-UMAP (Uniform Manifold Approximation and Projection) processing demonstrates significantly more distinct boundaries between clusters and structural patterns in the multitask setting (Fig. 5d). Together, joint learning across machine learning problem classes improves both quantitative performance as well as qualitative understanding of the source tasks.

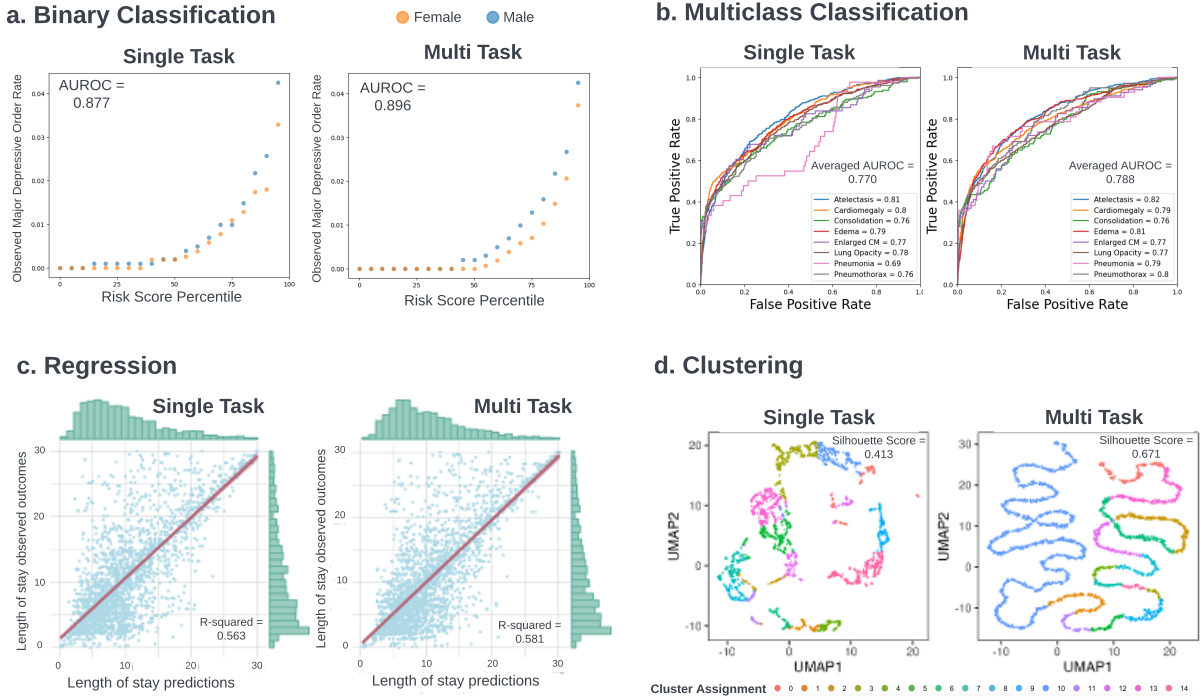

**Figure 5. MultiTask outperforms SingleTask across the four machine learning problem classes.**

## 6. Explainability

### 6.1 Task Interaction Measurement (TIM) Score Formulation

The M3H framework also provides explainability of task-dependency by computing a task interaction measurement (TIM) score, which measures how joint training of additional tasks affects the performance of the source task. It helps identify tasks that should be trained together to provide qualitative medical insights into potential connections between medical domains. Given $M$ as the number of all possible tasks, $S$ as a set of tasks that do not contain either task $i$ or task $j$, and $\tilde{f}_x(S \cup \{i\})$ as a function of the performance score of task $i$ given features $x$ and joint learning all tasks belonging to $S$ and task $i$, we define TIM as:

$$\delta_{ij} = \frac{1}{2^{M-2}} \sum_{S \subseteq \backslash \{i,j\}} \tilde{f}_x(S \cup \{i,j\}) - \tilde{f}_x(S \cup \{i\}))$$

As the number of all possible tasks grows, this score requires an exponentially increasing number of all potential task combinations of $S$. In practice, to avoid computational hurdles, we can either sample a subset of potential *S* to obtain an approximation of the true TIM score or restrict the number of task pair sizes to be small (i.e., smaller than 5).

**6.2 Task Interdependency Understanding**

We show in Fig. 6 that using the proposed task interaction measurement (TIM) score, we can quantify both the positive and negative contribution of additional tasks on a source task. Notably, consistent with previous findings, the additional joint learning of infectious diseases helps improve the forecast of length of stay (Rowell-Cunsolo 2018), and inflammatory bowel disease learning contributes to bipolar disorder risk prediction (Bernstein 2019). We compare multitask models of all pairwise task combinations of size 2 (restricted to a small number to ensure computational efficiency) against single-task models using only the source task across various medical domains. Each cell in the heatmap represents the rank of the TIM score between the source task (vertical axis) and target task (horizontal axis). The TIM scores of all target tasks are computed, ranked, and then reported for each source task. Target tasks with negative TIM scores are assigned negative ranks (colored blue), while those with positive TIM scores are assigned positive ranks (colored red). Diagonal cells are marked with zeros (colored white) since source and targets are the same tasks. We remark that the heatmap is not symmetric, underscoring that the direction of task interdependencies matters, as the effect of task A on task B may differ from the effect of task B on task A. This asymmetry highlights the complex nature of task relationships and their varying impacts depending on the direction of the interdependency and suggests that when designing multi-task models, it is important to clarify the rank of objectives when multiple tasks are jointly learned. Overall, the TIM score helps understand whether a particular task combination improves individual learning by sharing knowledge, impairs learning by competing between conflictive objectives, and can provide qualitative insights to better understand under-investigated medical outcome connections.

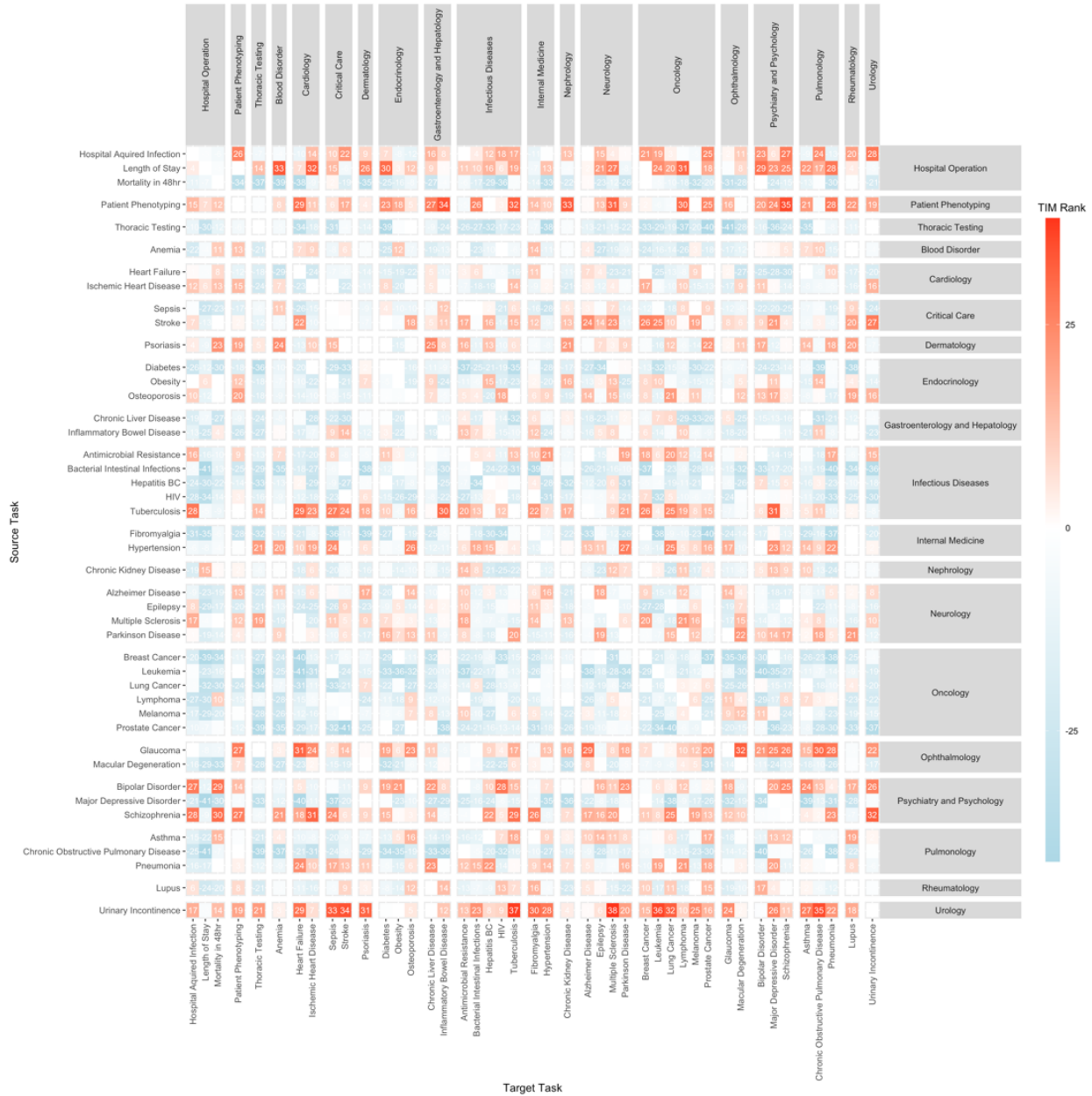

**Figure 6: Heatmap of TIM Scores Across Medical Tasks to Illustrate Task Interactions.**

## 7. Managerial Implications and Limitations

### 7.1 Implications to Practice

M3H can be readily adopted in production for hospital systems, especially in resource-constrained settings. By leveraging a modular architecture, M3H is adaptable to each system's specific patient cohorts, available data modalities, and targeted medical tasks of interest. This versatility facilitates user-defined modifications, replacements, and extensions, ensuring a tailored application in diverse hospital environments. M3H is also developed to be easily implementable with standard data storage or computational infrastructures available in most hospital settings. It has been packaged and tested on standard PCs as an executable software, and preliminary testing showed its feasibility in these local systems across both the Linux and Mac operating systems. Particularly in resource-constrained hospital systems, where information technology (IT) departments lack the capacity to manage huge-scale models, M3H offers a scalable alternative to democratize the use of such AI systems. Once validated, these prototypes can be implemented both in the clinical care delivery and the administrative operation management routines.

M3H also holds the potential to significantly improve operational efficiencies across areas, including patient flow forecast, wait time optimization, and cost predictions. For example, early accurate predictions of length of stay (LOS) and mortality enable physicians to better anticipate patient discharges to avoid flow congestions and manage patient wait time expectations; more accurate predictions of hospital-acquired infections allow hospitals to project potential reimbursement and billing complications more realistically with associated insurance payers. Done with appropriate associated analyses, M3H can lead to better operational efficiencies and reduce unnecessary hospital costs.

For patients, the most important benefit of the M3H is its ability to provide a deeper understanding of how multiple conditions can interact and affect each other. This insight not only improves early detection of single diseases but also equips patients with crucial information about their overall health. Consequently, patients can engage in more effective and targeted discussions with their healthcare providers, ensuring that the treatment for one condition does not adversely affect another, such as conflicting treatment or overuse of medications. Such integrated care could also reduce the financial burdens of patients by minimizing hospital visits and avoiding duplicating prescriptions or procedures, improving the overall care delivery experience.

**7.2 Limitations and Future Work**

There are a few important avenues that M3H can benefit from. First, M3H did not integrate the complete profile of data modalities and medical tasks available to HAIM-MM-MIMIC. For example, image segmentation and large language modeling (such as discharge note generation) can also be included in the framework as additional tasks. Furthermore, M3H should include other data modalities once they become available, such as omics (such as transcriptome profiling, simple nucleotide variation, and DNA methylation), as well as wearable device signals (such as lead 2 ECG data). Another important aspect of the M3H framework that should be investigated is its robustness towards data perturbation, as well as the numerical instability inherited by most deep learning architectures. Recent works on distributionally robust optimization can help prevent these issues from occurring by protecting against potential uncertainties introduced in the framework. Furthermore, designing what tasks to include for a multi-disease setting is a nonlinear, combinatorial problem that can be challenged by the curse of dimensionality as the number of possible combinations explodes. Some recent works have been done to explore Pareto optimal disease combinations, which is a promising direction to explore. Lastly, an interesting usage of the M3H framework is in connection with the predict-then-optimization literature. With more accurate performance in the prediction phase, it is possible that we can simultaneously improve multiple downstream optimization problems for better operational efficiency and recover analytical insights across medical departments. If these new directions are incorporated, we hope to improve the current capabilities of M3H to allow its wider applicability to even more healthcare applications.

# REFERENCES

Abadi M., Agarwal A., Barham P., Brevdo E., Chen Z., Citro C., Corrado G.S., et al. TensorFlow: Large-scale machine learning on heterogeneous systems, 2015.

Acosta, J. N., Falcone, G. J., Rajpurkar, P., & Topol, E. J. (2022). Multimodal biomedical AI. Nature Medicine, 28(9), 1773-1784. https://doi.org/10.1038/s41591-022-01981-2

Ahmed Z, Mohamed K, Zeeshan S, Dong X. Artificial intelligence with multi-functional machine learning platform development for better healthcare and precision medicine. Database (Oxford). 2020 Jan 1;2020:baaa010. doi: 10.1093/database/baaa010. PMID: 32185396; PMCID: PMC7078068.

Ahmed A., Xi R., Hou M., Shah S.A. and Hameed S., "Harnessing Big Data Analytics for Healthcare: A Comprehensive Review of Frameworks, Implications, Applications, and Impacts," in *IEEE Access*, vol. 11, pp. 112891-112928, 2023, doi: 10.1109/ACCESS.2023.3323574.

Alsentzer, E., Murphy, J. R., Boag, W., Weng, W., Jin, D., Naumann, T., & McDermott, M. B. (2019). Publicly Available Clinical BERT Embeddings. *ArXiv*. /abs/1904.03323

Alshmrani, G. M. M., Ni, Q., Jiang, R., Pervaiz, H., & Elshennawy, N. M. (2023). A deep learning architecture for multi-class lung diseases classification using chest X-ray (CXR) images. *Alexandria Engineering Journal*, *64*, 923-935. https://doi.org/10.1016/j.aej.2022.10.053

Anguita, D., Ghio, A., Oneto, L., Parra, X., Reyes-Ortiz, J.L. (2012). Human Activity Recognition on Smartphones Using a Multiclass Hardware-Friendly Support Vector Machine. In: Bravo, J., Hervás, R., Rodríguez, M. (eds) Ambient Assisted Living and Home Care. IWAAL 2012. Lecture Notes in Computer Science, vol 7657. Springer, Berlin, Heidelberg. https://doi.org/10.1007/978-3-642-35395-6_30

Apergi, L. A., Bjarnadóttir, M. V., Baras, J. S., & Golden, B. L. (2023). Cost Patterns of Multiple Chronic Conditions: A Novel Modeling Approach Using a Condition Hierarchy. In INFORMS Journal on Data Science. Institute for Operations Research and the Management Sciences (INFORMS). https://doi.org/10.1287/ijds.2022.0010

Baltrusaitis, T., Ahuja, C., & Morency, L.-P. (2019). Multimodal Machine Learning: A Survey and Taxonomy. IEEE Trans. Pattern Anal. Mach. Intell. 41, 2 (February 2019), 423–443. https://doi.org/10.1109/TPAMI.2018.2798607

Bentayeb, D., Lahrichi, N. & Rousseau, LM. On integrating patient appointment grids and technologist schedules in a radiology center. *Health Care Manag Sci* **26**, 62–78 (2023). https://doi.org/10.1007/s10729-022-09618-z

Bergman, D., Huang, T., Brooks, P., Lodi, A., & Raghunathan, A. U. (2021) JANOS: An Integrated Predictive and Prescriptive Modeling Framework. INFORMS Journal on Computing 34(2):807-816. https://doi-org.libproxy.mit.edu/10.1287/ijoc.2020.1023

Bertsimas, D., & Pauphilet, J. (2023). Hospital-Wide Inpatient Flow Optimization. In Management Science. Institute for Operations Research and the Management Sciences (INFORMS). https://doi.org/10.1287/mnsc.2023.4933

Bertsimas, D., Pauphilet, J., Stevens, J., & Tandon, M. (2022). Predicting Inpatient Flow at a Major Hospital Using Interpretable Analytics. In Manufacturing & Service Operations Management (Vol. 24, Issue 6, pp. 2809–2824). Institute for Operations Research and the Management Sciences (INFORMS). https://doi.org/10.1287/msom.2021.0971

Bish, D. R., Bish, E. K., & El Hajj, H. (2024). Disease Bundling or Specimen Bundling? Cost- and Capacity-Efficient Strategies for Multidisease Testing with Genetic Assays. In Manufacturing & Service Operations Management (Vol. 26, Issue 1, pp. 95–116). Institute for Operations Research and the Management Sciences (INFORMS). https://doi.org/10.1287/msom.2022.0296

Buergel, T., Steinfeldt, J., et al(2022). Metabolomic profiles predict individual multidisease outcomes. Nature Medicine, 28(11), 2309-2320. https://doi.org/10.1038/s41591-022-01980-3

Caruana, R., Lou, Y., Gehrke, J., Koch, P., Sturm, M., & Elhadad, N. 2015. Intelligible Models for HealthCare: Predicting Pneumonia Risk and Hospital 30-day Readmission. In Proceedings of the 21th ACM SIGKDD International Conference on Knowledge Discovery and Data Mining (KDD '15). Association for Computing Machinery, New York, NY, USA, 1721–1730. https://doi.org/10.1145/2783258.2788613

Bernstein N. C., Hitchon A. C., Walld R., Bolton M. J., et al. (2019), Increased Burden of Psychiatric Disorders in Inflammatory Bowel Disease, Inflammatory Bowel Diseases, Volume 25, Issue 2, February 2019, Pages 360–368, https://doi.org/10.1093/ibd/izy235

Chan, T. C. Y., Mahmood, R., O'Connor, D. L., Stone, D., et al. (2023). Got (Optimal) Milk? Pooling Donations in Human Milk Banks with Machine Learning and Optimization. In Manufacturing Service Operations Management. https://doi.org/10.1287/msom.2022.0455

Cohen, J. P., Viviano, J. D., Bertin, P., Morrison, P., Torabian, P., Guarrera, M., Lungren, M. P., Chaudhari, A., Brooks, R., Hashir, M., & Bertrand, H. (2021). TorchXRayVision: A library of chest X-ray datasets and models. *ArXiv*. /abs/2111.00595

Eyuboglu, S., Angus, G., Patel, B.N. *et al.* Multi-task weak supervision enables anatomically-resolved abnormality detection in whole-body FDG-PET/CT. *Nat Commun* 12, 1880 (2021). https://doi.org/10.1038/s41467-021-22018-1

Ferreiro A.L. et al. ,Gut microbiome composition may be an indicator of preclinical Alzheimer's disease.Sci. Transl. Med.15,eabo2984(2023).DOI:10.1126/scitranslmed.abo2984

Fu, S., Lai, H., Li, Q., Liu, Y., Zhang, J., Huang, J., Chen, X., Duan, C., Li, X., Wang, T., He, X., Yan, J., Lu, L., & Huang, M. (2021). Multi-task deep learning network to predict future macrovascular invasion in hepatocellular carcinoma. In eClinicalMedicine (Vol. 42, p. 101201). Elsevier BV. https://doi.org/10.1016/j.eclinm.2021.101201

Hajjar, A., & Alagoz, O. (2023). Personalized Disease Screening Decisions Considering a Chronic Condition. In Management Science (Vol. 69, Issue 1, pp. 260–282). https://doi.org/10.1287/mnsc.2022.4336

Haraty, R. A., Dimishkieh, M., & Masud, M. (2015). An Enhanced k-Means Clustering Algorithm for Pattern Discovery in Healthcare Data. *International Journal of Distributed Sensor Networks*. https://doi.org/10.1155/2015/615740

He, K., Gkioxari, G., Dollár, P., & Girshick, R.B. (2017). Mask R-CNN. *2017 IEEE International Conference on Computer Vision (ICCV)*, 2980-2988.

Hillerman, T., Souza, J. C. F., Reis, A. C. B., & Carvalho, R. N. (2017). Applying clustering and AHP methods for evaluating suspect healthcare claims. *Journal of Computational Science*, *19*, 97-111. https://doi.org/10.1016/j.jocs.2017.02.007

Hoang, L., Lee, S., Lee, E., & Kwon, K. (2022). Multiclass Skin Lesion Classification Using a Novel Lightweight Deep Learning Framework for Smart Healthcare. *Applied Sciences*, *12*(5), 2677. https://doi.org/10.3390/app12052677

Huang, S.-C., Pareek, A., Seyyedi, S., Banerjee, I. & Lungren, M. P. Fusion of medical imaging and electronic health records using deep learning: a systematic review and implementation guidelines. NPJ Dig. Med. 3, 1–9 (2020).

Ishigami J, Trevisan M, Xu H, Coresh J, Matsushita K, Carrero JJ. Estimated GFR and Hospital-Acquired Infections Following Major Surgery. Am J Kidney Dis. 2019 Jan;73(1):11-20. doi: 10.1053/j.ajkd.2018.06.029. Epub 2018 Sep 7. PMID: 30201545.

Jin, C., Yu, H., Ke, J. *et al.* Predicting treatment response from longitudinal images using multi-task deep learning. *Nat Commun* **12**, 1851 (2021). https://doi.org/10.1038/s41467-021-22188-y

Johnson, A. E., Bulgarelli, L., Shen, L., Gayles, A., Shammout, A., Horng, S., Pollard, T. J., Hao, S., Moody, B., Gow, B., Lehman, L., Celi, L. A., & Mark, R. G. (2023). MIMIC-IV, a freely accessible electronic health record dataset. *Scientific Data*, *10*(1), 1-9. https://doi.org/10.1038/s41597-022-01899-x

Kim, K., Mehrotra, S. (2015) A Two-Stage Stochastic Integer Programming Approach to Integrated Staffing and Scheduling with Application to Nurse Management. Operations Research 63(6):1431-1451. https://doi-org.libproxy.mit.edu/10.1287/opre.2015.1421

Kraus, M., Feuerriegel, S., & Saar-Tsechansky, M. (2024). Data-Driven Allocation of Preventive Care with Application to Diabetes Mellitus Type II. In Manufacturing & Service Operations Management (Vol. 26, Issue 1, pp. 137–153). Institute for Operations Research and the Management Sciences (INFORMS). https://doi.org/10.1287/msom.2021.0251

Lan Y., Chandrasekaran A., Goradia D., Walker D. (2022) Collaboration Structures in Integrated Healthcare Delivery Systems: An Exploratory Study of Accountable Care Organizations. Manufacturing &

Service Operations Management 24(3):1796-1820. https://doi-org.libproxy.mit.edu/10.1287/msom.2021.1038

Lee, M.H., Kim, N., Yoo, J. *et al.* Multitask fMRI and machine learning approach improve prediction of differential brain activity pattern in patients with insomnia disorder. *Sci Rep* **11**, 9402 (2021). https://doi.org/10.1038/s41598-021-88845-w

Liu, S., Johns, E., & Davison, A. J. (2018). End-to-End Multi-Task Learning with Attention. ArXiv. /abs/1803.10704

Misra I., Shrivastava A., Gupta A., and Hebert M.. Cross-stitch networks for multi-task learning. In Proceedings of the IEEE Conference on Computer Vision and Pattern Recognition, pages 3994–4003, 2016.

Na, L., Carballo, K. V., Pauphilet, J., Kombert, D., Castiglione, et al. (2023). Patient Outcome Predictions Improve Operations at a Large Hospital Network. ArXiv. /abs/2305.15629

Peasah SK, McKay NL, Harman JS, Al-Amin M, Cook RL. Medicare non-payment of hospital-acquired infections: infection rates three years post implementation. Medicare Medicaid Res Rev. 2013 Sep 25;3(3):mmrr.003.03.a08. doi: 10.5600/mmrr.003.03.a08. PMID: 24753974; PMCID: PMC3983733.

Pogorelov K., Randel K.R., Griwodz C., Eskeland S.L., Lange T., Johansen D., Spampinato C., Dang-Nguyen D-T, Lux M., Schmidt P. T., Riegler M., and Halvorsen P.. 2017. KVASIR: A Multi-Class Image Dataset for Computer Aided Gastrointestinal Disease Detection. https://doi.org/10.1145/3193289

Radford, Alec, Jeff Wu, Rewon Child, David Luan, Dario Amodei and Ilya Sutskever. "Language Models are Unsupervised Multitask Learners." (2019).

Rawshani A, Rawshani A, Franzén S, Sattar N, Eliasson B, Svensson AM, Zethelius B, Miftaraj M, McGuire DK, Rosengren A, Gudbjörnsdottir S. Risk Factors, Mortality, and Cardiovascular Outcomes in Patients with Type 2 Diabetes. N Engl J Med. 2018 Aug

Reed, S., Zolna, K., Parisotto, E., Colmenarejo, S. G., Novikov, A., Gimenez, M., Sulsky, Y., Kay, J., Springenberg, J. T., Eccles, T., Bruce, J., Razavi, A., Edwards, A., Heess, N., Chen, Y., Hadsell, R., Vinyals, O., Bordbar, M., & De Freitas, N. (2022). A Generalist Agent. *ArXiv*. /abs/2205.06175

Reid KW, Vittinghoff E, Kushel MB. Association between the level of housing instability, economic standing and health care access: a meta-regression. J Health Care Poor Underserved. 2008 Nov;19(4):1212-28. doi: 10.1353/hpu.0.0068. PMID: 19029747.

Ren, S., He, K., Girshick, R., & Sun, J. (2015). Faster r-cnn: Towards real-time object detection with region proposal networks. *Advances in Neural Information Processing Systems*, *28*.

Rowell-Cunsolo TL, Liu J, Shen Y, Britton A, Larson E. The impact of HIV diagnosis on length of hospital stay in New York City, NY, USA. AIDS Care. 2018 May;30(5):591-595. doi: 10.1080/09540121.2018.1425362. Epub 2018 Jan 17. PMID: 29338331; PMCID: PMC5860957.

Salari N, Khazaie H, Hosseinian-Far A, Khaledi-Paveh B, Kazeminia M, Mohammadi M, Shohaimi S, Daneshkhah A, Eskandari S. The prevalence of stress, anxiety and depression within front-line healthcare workers caring for COVID-19 patients: a systematic review and meta-regression. Hum Resour Health. 2020 Dec 17;18(1):100. doi: 10.1186/s12960-020-00544-1. PMID: 33334335; PMCID: PMC7745176.

Soenksen, L. R., Ma, Y., Zeng, C., Boussioux, L., Villalobos Carballo, K., Na, L., Wiberg, H. M., Li, M. L., Fuentes, I., & Bertsimas, D. (2022). Integrated multimodal artificial intelligence framework for healthcare applications. *Npj Digital Medicine*, *5*(1), 1-10. https://doi.org/10.1038/s41746-022-00689-4

Tang, X., Zhang, J., He, Y. *et al.* Explainable multi-task learning for multi-modality biological data analysis. *Nat Commun* **14**, 2546 (2023). https://doi.org/10.1038/s41467-023-37477-x

Topol, E. Deep medicine: how artificial intelligence can make healthcare human again. (Hachette UK, 2019).

Torres-Soto, J., Ashley, E.A. Multi-task deep learning for cardiac rhythm detection in wearable devices. *npj Digit. Med.* **3**, 116 (2020). https://doi.org/10.1038/s41746-020-00320-4

Tseng, V.WS., Sano, A., Ben-Zeev, D. *et al.* Using behavioral rhythms and multi-task learning to predict fine-grained symptoms of schizophrenia. *Sci Rep* **10**, 15100 (2020). https://doi.org/10.1038/s41598-020-71689-1

Vaswani, A., Shazeer, N., Parmar, N., Uszkoreit, J., Jones, L., Gomez, A. N., Kaiser, Ł., & Polosukhin, I. 2017. Attention is all you need. In Proceedings of the 31st International Conference on Neural Information Processing Systems (NIPS'17). Curran Associates Inc., Red Hook, NY, USA, 6000–6010.

Wang, G., Li, J., & Hopp, W. J. (2023). Personalized Healthcare Outcome Analysis of Cardiovascular Surgical Procedures. In Manufacturing & Service Operations Management (Vol. 25, Issue 4, pp. 1567–1584). Institute for Operations Research and the Management Sciences (INFORMS). https://doi.org/10.1287/msom.2023.1227

Wang, X., Cheng, Y., Yang, Y. *et al.* Multitask joint strategies of self-supervised representation learning on biomedical networks for drug discovery. *Nat Mach Intell* **5**, 445–456 (2023). https://doi.org/10.1038/s42256-023-00640-6

Yang, Y., Goldhaber-Fiebert, J. D., & Wein, L. M. (2013). Analyzing Screening Policies for Childhood Obesity. In Management Science (Vol. 59, Issue 4, pp. 782–795). https://doi.org/10.1287/mnsc.1120.1587

Yu, K., Beam, A. L., & Kohane, I. S. (2018). Artificial intelligence in healthcare. Nature Biomedical Engineering, 2(10), 719-731. https://doi.org/10.1038/s41551-018-0305-z

Zhao, Y., Zhuang, Z., Li, Y., Xiao, et al. (2024). Elevated blood remnant cholesterol and triglycerides are causally related to the risks of cardiometabolic multimorbidity. Nature Communications, 15(1), 1-9. https://doi.org/10.1038/s41467-024-46686-x